\documentclass[conference]{IEEEtran}
\IEEEoverridecommandlockouts
\usepackage{cite}
\usepackage{amsmath,amssymb,amsfonts}
\usepackage{algorithmic}
\usepackage{graphicx}
\usepackage{textcomp}
\usepackage{xcolor}
\usepackage{multirow}
\usepackage{booktabs}
\usepackage{makecell}
\def\BibTeX{{\rm B\kern-.05em{\sc i\kern-.025em b}\kern-.08em
    T\kern-.1667em\lower.7ex\hbox{E}\kern-.125emX}}
\begin{document}

\title{Sentence Simplification Using Paraphrase Corpus for Initialization  }

\author{\IEEEauthorblockN{Kang Liu, Jipeng Qiang}
\IEEEauthorblockA{\textit{Department of Computer Science, Yangzhou University} \\
Jiangsu, China \\
yzuNLPlk@163.com, jpqiang@yzu.edu.cn}

}

\maketitle

\begin{abstract}

Neural sentence simplification method based on sequence-to-sequence framework has become the mainstream method for sentence simplification (SS) task. Unfortunately, these methods are currently limited by the scarcity of parallel SS corpus. In this paper, we focus on how to reduce the dependence on parallel corpus by leveraging a careful initialization for neural SS methods from paraphrase corpus. Our work is motivated by the following two findings: (1) Paraphrase corpus includes a large proportion of sentence pairs belonging to SS corpus. (2) We can construct large-scale pseudo parallel SS data by keeping these sentence pairs with a higher complexity difference. Therefore, we propose two strategies to initialize neural SS methods using paraphrase corpus.  We train three different neural SS methods with our initialization, which can obtain substantial improvements on the available WikiLarge data compared with themselves without initialization. 

\end{abstract}


\section{Introduction}

The goal of sentence simplification (SS) task is to rephrase a sentence into a form that is easier to read and understand, while still retaining the semantic meaning, which can help people with reading difficulties such as non-native speakers\cite{paetzold2016lexical,qiang2021chinese}, dyslexia\cite{rello2013simplify} or autism\cite{evans2014evaluation}. Second language learners\cite{xia-etal-2016-text} and people with low literacy\cite{watanabe2009facilita} can also benefit from it.

Since the 2010 year, SS task have been addressed as a monolingual machine translation problem translating from complex sentences to simplified sentences. Existing SS methods have changed from statistical sentence simplification methods \cite{zhu2010monolingual,narayan2014hybrid,xu2016optimizing} to neural sentence simplification methods \cite{zhang-lapata-2017-sentence,surya2018unsupervised,martin2020multilingual,agrawal-etal-2021-non}. Neural sentence simplification methods adopt sequence to sequence (Seq2Seq) models. Seq2Seq models work very well only when provided with a massive parallel corpus of complex and simplified sentences. Unfortunately, these approaches are currently limited by the scarcity of parallel corpus. For example, the biggest and widely-used SS training dataset WikiLarge \cite{zhang-lapata-2017-sentence} is composed of 296,402 sentence pairs, which align sentences from the 'ordinary' English Wikipedia and the 'simple' English Wikipedia. WikiLarge has been criticized recently \cite{xu2015problems,stajner-2021-automatic} because they contain a large proportion of noise data, which leads to systems that generalize poorly. Some work \cite{kajiwara2018text,martin2020multilingual,lu2021unsupervised,qiang2023parals} foucs on unsupervised SS method for alleviating the need for SS supervised corpora. In this paper, we focus on how to reduce the dependence on parallel corpus by leveraging a careful initialization for neural SS methods.

There are large-scale paraphrase datasets \cite{Wieting2017,hu2019parabank} for paraphrase generation whose aim is to generate an output sentence that preserves the meaning of the input sentence but contains variations in word choice and grammar. Comparing with paraphrase dataset, SS dataset highlights that the two sentences of each sentence pair should have difference in text complexity levels. We found that there are a large proportion of sentence pairs in paraphrase dataset that satisfy the expectations of SS task. For example, paraphrase dataset ParaBank \cite{hu2019parabank} was created automatically from bilingual text by pivoting over the non-English language using neural machine translation (NMT) models. NMT models usually tend to generate more high-frequency tokens and less low-frequency tokens \cite{Jiang2019,gu-etal-2020-token}. Considering that the higher the word frequency, the simpler the word is, this phenomenon could be beneficial to SS task. Table 1 shows two sentence pairs from paraphrase corpus ParaBank. We can see that the translated target sentence is easier than the source sentence.

\begin{table}[h!]
\begin{center}
\textbf{Table 1}~~ Two examples in ParaBank paraphrase corpus\\
\setlength{\tabcolsep}{1mm}{
\begin{tabular}{l|l} \toprule
\multirow{2}{*}{\textbf{Source}} &\makecell[l]{This \textbf{proposal} will be \textbf{communicated} to the trader 's } \\ 
&  creditors.  \\ \hline
\textbf{Target}&\makecell[l]{This \textbf{plan} will be \textbf{sent} to the trader 's creditors} \\\hline\hline
\multirow{2}{*}{\textbf{Source}} &\makecell[l]{It would be \textbf{prudent} for you not to be \textbf{deceived} by }\\ 
& your masquerade .  \\ \hline
\multirow{2}{*}{\textbf{Target}} &\makecell[l]{It would be \textbf{wise} for you not to be \textbf{fooled} by your } \\
&  own masquerade .  \\ \hline
\bottomrule
\end{tabular}}
\end{center}
\end{table}

In this paper, we will try to utilize paraphrase corpus to initialize neural SS methods, and then fine-tune these methods on real SS dataset. Specifically, we design two strategies for initialization. (1) We directly utilize the whole paraphrase corpus to train an initial SS method. (2) Considering many sentences pairs in paraphrase corpus cannot satisfy the expectations of SS task, we only select these sentence pairs with a higher complexity difference using text readability formula (Flesch reading ease score \cite{kincaid1975derivation}), which is designed to indicate how difficult a sentence is to understand. Experimental results show that neural SS methods based on our initialization outperform themselves without initialization. 

The following sections are organized as follows: Section 2 describes the related work; Section 3 presents how to initialize neural SS methods; Section 4 shows the experimental results; Section 5 summarizes the paper.

\section{Related Work}

\subsection{Sentence Simplification}

Automatic SS is a complicated natural language processing (NLP) task, which consists of lexical and syntactic simplification levels. It has attracted much attention recently as it could make texts more accessible to wider audiences, and used as a pre-processing step, improve performances of various NLP tasks and systems.
Usually, hand-crafted, supervised, and unsupervised methods based on resources like English Wikipedia and Simple English Wikipedia (EW-SEW) \cite{coster2011simple} are utilized for extracting simplification rules. It is very easy to mix up the automatic TS task and the automatic summarization task \cite{zhu2010monolingual,rush2015neural}. TS is different from text summarization as the focus of text summarization is to reduce the length and redundant content.

At the lexical level, lexical simplification often substitutes difficult words using more common words, which only require a large corpus of regular text to obtain word embeddings to get words similar to the complex word \cite{glavavs2015simplifying,qiang2020lsbert,qiang2023natural,qiang2021lsbert}. Woodsend and Lapata \cite{woodsend2011learning} presented a data-driven model based on a quasi-synchronous grammar, a formalism that can naturally capture structural mismatches and complex rewrite operations. Wubben et al. \cite{wubben2012} proposed a phrase-based machine translation (PBMT) model that is trained on ordinary-simplified sentence pairs. Xu et al. \cite{xu2016optimizing} proposed a syntax-based machine translation model using simplification-specific objective functions and features to encourage simpler output. 

Neural machine translation has shown to produce state-of-the-art results \cite{bahdanau2014neural,artetxe-etal-2018-unsupervised,gu-etal-2020-token}, which are based on sequence-to-sequence (Seq2Seq) architecture. In recent years, many neural SS models based on Seq2Seq are proposed and achieve good results \cite{nisioi2017exploring,zhang-lapata-2017-sentence,surya2018unsupervised,martin2020multilingual,agrawal-etal-2021-non,qiang2019heterogeneous}. The main limitation of the aforementioned neural SS models depended on the parallel ordinary-simplified sentence pairs \cite{stajner-2021-automatic}. Because ordinary-simplified sentence pairs are expensive and time-consuming to build, the available largest data is WikiLarge \cite{zhang-lapata-2017-sentence} that only has 296,402 sentence pairs. The dataset is insufficiency for neural SS model if we want to they can obtain the best parameters. Considering paraphrase corpus includes a large number of sentence pairs that satisfy the expectations of SS task. In this paper, we investigate the use of paraphrase data for text simplification. We are the first to show that we can effectively adapt paraphrase data for SS task.

\subsection{Unsupervised Sentence Simplification}

To overcome the scarcity of parallel SS corpus, unsupervised SS methods without using any parallel corpus have attracted much attention. Existing unsupervised SS methods can be divided into two classifications. The first scheme focuses on how to design an unsupervised SS method, and the second scheme concentrates on how to build a parallel SS corpus. 

\cite{2015Unsupervised} and \cite{kumar-etal-2020-iterative} are the pipeline-based unsupervised framework, where the pipeline of Narayan and Gardent is composed of lexical simplification, sentence splitting, and phrase deletion, the pipeline of Kumar et al. includes deletion, reordering, and lexical simplification. \cite{surya-etal-2019-unsupervised} proposed an unsupervised neural text simplification based on a shared encoder and two decoders, which only learn the neural network parameters from simple sentences set and complex sentences set. In other languages, there are unsupervised statistical machine translations for Japanese \cite{katsuta2019improving} and back-translation in Spanish and Italian \cite{palmero2019neural}. The performance of the above unsupervised SS methods is however often below their supervised counterparts.

Some work \cite{kajiwara2018text,martin2020multilingual} constructed SS corpora by searching the most similar sentences using sentence embedding modeling, and train SS methods using the constructed SS corpora. \cite{kajiwara2018text} calculated the similarity between the sentences from English Wikipedia by Word Mover's distance \cite{kusner2015word}. \cite{martin2020multilingual} adopted multilingual sentence embedding modeling LASER \cite{artetxe-etal-2018-unsupervised} to calculate the similarity between the sentences from 1 billion sentences from CCNET \cite{Wenzek2019}. Since the aim of the two works is to find the most similar sentences from a large corpus, they cannot guarantee that the aligned sentences preserve the same meanings. Lv et al. \cite{lu-etal-2021-unsupervised-method} construct large-scale pseudo parallel SS data by taking the pair of the source sentences of translation corpus and the translations of their references in a bridge language. 

\subsection{ Paraphrase Mining }

Some work has focused on generating paraphrase corpus for neural machine translation (NMT) systems using back-translation, where back-translation \cite{sennrich2015improving} is a technique widely used in NMT to enhance the target monolingual data during the training process. Specifically, the back-translation technique is used by translating the non-English side of bitexts back to English\cite{wieting2017learning} and pairing translations with the references. Two large paraphrase corpora (PARANMT\cite{wieting2017paranmt} and PARABANK\cite{hu2019parabank}) are built based on this idea, and has been proven to have great potential in different translation-core tasks. Round-trip translation is also used in mining paraphrases \cite{mallinson2017paraphrasing} by translating sentences into another language then translating the result back into the original language. Similar to machine translation, back-translation is used to improve the performance of neural SS methods \cite{katsuta2019improving,palmero2019neural,qiang2021unsupervised}. \cite{mehta2020simplifythentranslate} trained a paraphrasing model by generating a paraphrase corpus using back-translation, which is used to preprocess source sentences of the low-resource language pairs before feeding into the NMT system.

The above work for building a large paraphrase corpus is to serve for NMT and other tasks, which is not fit for SS task. The difference of sentence complexity between the original sentence and the translated sentence for each sentence pair has not been taken into consideration, which is vitally important for SS task. Therefore, we focus on how to build a sentence simplification corpus, instead of a paraphrase corpus.

\section{Method}

In this section, we will present how to utilize paraphrase corpus to initialize neural SS models. 

\subsection{Relation between Paraphrase Corpus and SS Corpus}

Some work has focused on generating paraphrase corpus for neural machine translation (NMT) systems using back-translation technique, where back-translation \cite{sennrich-etal-2016-improving} is a technique widely used in NMT to enhance the target monolingual data during the training process. Specifically, the back-translation technique is used by translating the non-English side of bitexts back to English \cite{wieting-gimpel-2018-paranmt} and pairing translations with the references. We can see that the two sentences of each sentence pair of paraphrase corpus should preserve the same meaning.

\textbf{Hypothesis 1: SS corpus can be regarded as a subset of paraphrase corpus.} Based on the definition of SS task, SS corpus should satisfy the following two requirements: (1) The two sentences of each sentence pair should convey the same meaning. (2) The two sentences of each sentence pair should have difference in text complexity levels. Paraphrase only needs to satisfy the first requirement, and SS corpus needs to satisfy both the requirements. 

\textbf{Hypothesis 2: Paraphrase corpus includes a large proportion of sentence pairs belonging to SS corpus.} Neural machine translation model usually tends to generate more high-frequency tokens and less low-frequency tokens \cite{Jiang2019,gu-etal-2020-token}. The frequency of words is one of the most popular choices by sentence simplification \cite{qiang2020lsbert,qiang2021unsupervised}. In general, the higher the frequency, the easier the word. Many empirical results supported the hypothesis, as shown in Table 1.

\subsection{Our Initialization Strategy}

We provide two strategies to initialize neural SS models.

\textbf{(1) First Initialization Strategy:} Based on Hypothesis 2, we directly utilize paraphrase corpus to train initial neural SS modeling. Here, we choose ParaBank as our using paraphrase corpus.  Due to the memory size of our computer, we only randomly choose 2 million sentence pairs from ParaBank \cite{hu2019parabank}. Finally, we train neural SS modeling on real SS corpus.

\begin{figure*}
    \resizebox{\textwidth}{!}{
    \centering
    \includegraphics{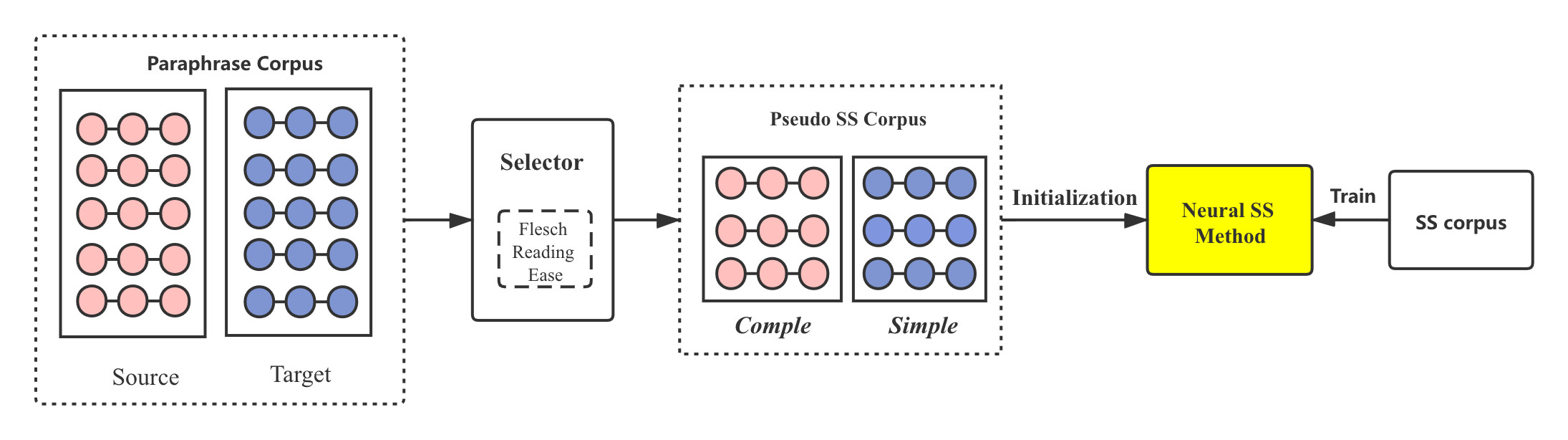}
    }
    \caption{The overview of our approach for training. A pseudo SS corpus is synthesized by selecting these complex-simple sentence pairs with a higher complexity difference. Then, we first train neural SS method using the pseudo SS corpus, and train neural SS method using the real SS corpus. }
    \label{fig:approach}
\end{figure*}

\textbf{(2) Second Initialization Strategy:} Our second initialization strategy is shown in Figure \ref{fig:approach}. Different from the first one, we only select these sentence pairs from paraphrase corpus that have difference in text complexity levels.  We measure the difference of text complexity using Flesch reading ease score (FRES) \cite{kincaid1975derivation}, which is designed to indicate how difficult a sentence is to understand, and is widely used to evaluate the performance of SS. FRES proposed in 1975 is a classical formula in the field of text assessment, whose coefficients are set by linguists. It is based on text features such as the average sentence length and the average number of syllables per word. A higher score indicates that the sentence is simpler to read. FRES grades the text from 0 to 100. The higher scores indicate the sentences are easier to read. As usual, the difference of one school grade level in FRES is 10, e.g., 5th grade (100.00-90.00) and 6th grade (90.0-80.0). The formula of FRES is,

\begin{equation}
\resizebox{.89\hsize}{!}{
$
    206.835-1.015\left(\frac{\text {\# words }}{\text {\# sentences }}\right)-84.6\left(\frac{\text {\# syllables }}{\text {\# words }}\right)
$
}
\label{sec:fres}
\end{equation}

To ensure simplicity, we only keep the sentence pairs with a FRES difference higher than a threshold $h_{\textit{FRES}}$. In our experiments, we set $h_{\textit{FRES}}$ = 10.0, where $h_{\textit{FRES}}=10.0$ means that for each sentence pair, the simplified version should be at least one school level simpler than its complex counterpart.

After obtaining pseudo SS corpus, we first initialize neural SS method using pseudo SS corpus, and train neural SS method on real SS corpus. 

\textbf{(3) Statistics of our choosing paraphrase corpora:}

We report the statistics of our choosing paraphrase corpora in Table 2. Here, we report the statistics of real SS corpus WikiLarge for a comparison. WikiLarge is the most popular and wildly used SS corpus. Because the SS task is a paraphrase generation task using easier words, the length of the complex sentence and the simple sentence are roughly the same, and the size of the vocabulary in the simple sentence set should be smaller than the complex sentence set. In contrast to the paraphrase corpora, the length of the complex sentence in WikiLarge is longer than the simple sentence, because it focuses on the deletion of content.

\begin{table}[h]
\centering\small

\textbf{Table 2}~~ Statistics of our choosing paraphrase corpora in two strategies compared with WikiLarge. Avg(complex) and Avg(simple) are the average numbers of words in the complex sentences and the simpler sentences, respectively. \\

\resizebox{\columnwidth}{!}{
\begin{tabular}{l|c|cc} \toprule
& \textbf{WikiLarge} & \textbf{First} & \textbf{Second } \\\midrule
\textbf{Vocab(complex)} & 169,349 & 282,279 & 96,524 \\
\textbf{Vocab(simple)} & 135,607 & 245,447 & 92,156 \\ \midrule
\textbf{Avg(complex)} & 21.93 & 12.04 & 10.49 \\
 \textbf{Avg(simple)} & 16.14 & 12.65 & 11.31 \\ \midrule
 \textbf{Total pairs} & 296,402 & 2,000,000 & 321,900 \\ \bottomrule
\end{tabular}
}

\end{table}

\section{Experinments}

\subsection{Experinmental setup}

\textbf{Neural SS methods:} To validate that our two initialization strategies (First and Second) are effective for different neural SS methods, we apply our two initialization on the following three methods:
\begin{itemize}
\item \textbf{LSTM} that is composed of RNN network and soft attention layer. 
\item \textbf{Transformer} that is based solely on attention mechanisms.
\item \textbf{Bart}\footnote{https://dl.fbaipublicfiles.com/fairseq/models/bart.base.tar.gz} that is sequence-to-sequence model trained with denoising as pretraining objective
\end{itemize}

We implement the above three methods via the opensource toolkit fairseq\cite{ott-etal-2019-fairseq}. We adopt the Adam optimizer with $ \beta_{1}=0.9 $, $ \beta_{2}=0.98 $, $ \epsilon = 10^{-8} $ and Dropout is set 0.3 for the three methods. The initial learning rate are set to $1 \times 10^{-4}$, $1 \times 10^{-4}$, $lr=1 \times 10^{-5}$ for LSTM-based, Transformer-based and BART-based models, respectively.

\textbf{Evaluation Dataset:} We select WikiLarge as the training SS corpus. For evaluating neural SS methods, we select TurkCorpus \cite{xu2016optimizing} as our evaluation benchmark dataset. The corpus consists of 2000 valid sentences and 359 test sentences. In TurkCorpus, each complex sentence has 8 kinds of simplification for reference. 

\textbf{Evaluation Metrics:} SARI\cite{xu2016optimizing} is the main metric to evaluate text simplification models, which calculates the arithmetic mean of the $n$-gram F1 scores of three operations (keeping, adding, and deleting) through comparing the generated sentences to multiple simplification references and the original sentences. A Higher SARI score means better simplification performance.  We use SS evaluation tools Easse\cite{alva-manchego-etal-2019-easse} to calculate the SARI metric.

\subsection{Experinmental Results}
\begin{table}[h!]
\begin{center}
\textbf{Table 3}~~ The evaluation result of the experiments. \\
\setlength{\tabcolsep}{6mm}{
\begin{tabular}{llc} \toprule
Model  & Condition &SARI  \\ \hline
\multirow{3}{*}{LSTM}&-&35.77   \\
&First  & 36.25  \\
&Second  &\textbf{36.45}    \\ \toprule
\multirow{3}{*}{Transformer}&-  & 37.29   \\
&First & 37.64  \\ 
&Second  & \textbf{38.17}   \\\toprule
\multirow{3}{*}{Bart}&-  & 38.03  \\
&First & 38.29 \\
&Second & \textbf{38.77}  \\
\bottomrule
\end{tabular}}
\end{center}
\end{table}

The final evaluation results are shown in Table 3. We can see that the three neural SS methods (LSTM, Transformer and Bart) with our First initialization strategy outperform themselves without initialization. The results indicate that our first initialization strategy is effective for neural SS methods. As we expected, the Second initialization method with a selector indeed further improves the performance of neural SS methods. With a selector, our paraphrase corpus becomes more suitable for SS task. The selector makes SARI score get 0.68 improved for LSTM, 0.88 improved for Transformer, 0.74 improved for Bart-based compared with themselves without initialization. From the simplified sentences, we found that Second also improves the readability of simplification results in varying degrees compared with the First initialization method without a selector. This indicates that the noise sentences in the paraphrase really harm the model training through the First initialization method. We can conclude that Second is a more reasonable and better method.

\begin{table}[h!]
\begin{center}
\textbf{Table 4}~~ The examples of simplified results generated by Transformer with the second initialization.\\
\setlength{\tabcolsep}{2mm}{
\begin{tabular}{llll} \toprule
Complex&\makecell[l]{it was \textbf{originally} thought that the debris thrown \\ up by the collision filled in the smaller craters .}\\
Reference&\makecell[l]{it was \textbf{first} thought that the debris thrown  \\ up by the collis-ion filled in the smaller craters .} \\
Transformer&\makecell[l]{it was \textbf{originally} thought that the debris thrown \\ up by coll-ision filled in the smaller craters .} \\
Second&\makecell[l]{it was \textbf{first} thought that the debris thrown \\ up by the collis-ion filled in the smaller craters .} \\\bottomrule
Complex&\makecell[l]{both names became defunct in 2007 when they were  \\ \textbf{merged} into the national museum of scotland .}\\
Reference&\makecell[l]{both names \textbf{merged with each other in 2007} to \\ \textbf{become}the national museum of scotland .} \\
Transformer&\makecell[l]{both names became defunct in 2007 when they were \\ \textbf{merged} into the national museum of scotland .} \\
Second&\makecell[l]{both names became defunct in 2007 when they were \\ \textbf{joined} into the national museum of scotland .} \\\bottomrule
\end{tabular}}
\end{center}
\end{table}

Table 4 shows the examples of the simplification result generated by Transformer without initialization and Transformer with our second initialization method. In the first example, we can find that our proposed method replaces 'originally' with 'first' which is the same as the reference while Transformer only repeats the original sentence. In the second example, our method replaces 'merged' with 'joined' while Transformer still repeats. This indicates that our initialization method can make more simplification with word replacement compared with the baseline method.

\section{Conclusions}
Considering the relationship between paraphrase corpus and SS corpus, we propose two strategies to initialize neural sentence simplification (SS) model using paraphrase corpus. Experimental results verify that neural SS methods without our initialization outperform themselves without initialization. In this paper, we use a small version of the paraphrase corpus. In future work, we can use a bigger paraphrase corpus. In addition to that, we can also build a new paraphrase corpus with different kinds of selectors not just FRES selector.

\bibliographystyle{ieeetr}
\bibliography{PAKDD2021.bib}
\end{document}